%% file: main.tex
\documentclass[conference]{IEEEtran}
\IEEEoverridecommandlockouts
\usepackage{cite}
\usepackage{amsmath,amssymb,amsfonts,amsthm}
\usepackage{algorithmic}
\usepackage{graphicx}
\usepackage{caption}
\usepackage{subcaption}
\usepackage{textcomp}
\usepackage{xcolor}

\theoremstyle{definition}
\newtheorem{definition}{Definition}[section]
\newcommand{\norm}[1]{\left\lVert#1\right\rVert}
\newcommand*\numcircledmod[1]{\raisebox{.5pt}{\textcircled{\raisebox{-.9pt} {#1}}}}

\def\BibTeX{{\rm B\kern-.05em{\sc i\kern-.025em b}\kern-.08em
    T\kern-.1667em\lower.7ex\hbox{E}\kern-.125emX}}
\begin{document}

\title{%MEMBERSHIP PRIVACY FOR IMAGE TRANSLATION MODELS VIA ADVERSARIAL KNOWLEDGE DISTILLATION
Membership Privacy Protection for Image Translation Models via Adversarial Knowledge Distillation
}

 \author{\IEEEauthorblockN{1\textsuperscript{st} Saeed Ranjbar Alvar$^*$}
 \IEEEauthorblockA{%\textit{dept. name of organization (of Aff.)} \\
 \textit{Simon Fraser University}\\
 %email address or ORCID
 }
 \and
 \IEEEauthorblockN{2\textsuperscript{nd} Lanjun Wang }
 \IEEEauthorblockA{%\textit{dept. name of organization (of Aff.)} \\
 \textit{Tianjin University}\\
 %email address or ORCID
 }
 \and
 \IEEEauthorblockN{3\textsuperscript{rd} Jian Pei }
 \IEEEauthorblockA{%\textit{dept. name of organization (of Aff.)} \\
 \textit{Simon Fraser University}\\
 %email address or ORCID
 }
 \and
 \IEEEauthorblockN{4\textsuperscript{th} Yong Zhang }
 \IEEEauthorblockA{%\textit{dept. name of organization (of Aff.)} \\
 \textit{Huawei Canada Technologies Co. Ltd.}\\
 %email address or ORCID
 }
% \and
% \and
}

\maketitle

\begin{abstract}
% This document is a model and instructions for \LaTeX.
% This and the IEEEtran.cls file define the components of your paper [title, text, heads, etc.]. *CRITICAL: Do Not Use Symbols, Special Characters, Footnotes, 
% or Math in Paper Title or Abstract.
Image-to-image translation models are shown to be vulnerable to the Membership Inference Attack (MIA), in which the adversary's goal is to identify whether a sample is used to train the model or not. With daily increasing applications based on image-to-image translation models, it is crucial to protect the privacy of  these models against MIAs. 

We propose adversarial knowledge distillation (AKD) as a defense method against MIAs for image-to-image translation models. The proposed method protects the privacy of the training samples by improving the generalizability of the model. 
We conduct experiments on the image-to-image translation models and show that AKD achieves the state-of-the-art utility-privacy tradeoff by reducing the attack performance up to 38.9\% compared with the regular training model at the cost of a slight drop in the quality of the generated output images. 
The experimental results also indicate that the models trained by AKD generalize better than the regular training models.
Furthermore, compared with existing defense methods, the results show that at the same privacy protection level, image translation models trained by AKD generate outputs with higher quality; while at the same quality of outputs, AKD enhances the privacy protection over 30\%.
%achieves a much higher privacy protection level. 
 %illustrate 
%a differentially private model and the models trained with earlier knowledge-distillation based defence proposed for image classification models.
%achieves better generalizability than trained by vanilla training.}
\end{abstract}

\let\thefootnote\relax\footnotetext{* This work was conducted while the first author was doing internship at Huawei Canada Technologies Co. Ltd.. }

\input{Sections/01_intro_Lanjun}
\input{Sections/02_preliminary}

\input{Sections/03_proposed}
\input{Sections/04_exp_setup}
\input{Sections/05_exp}

\input{Sections/06_relatedwork}
\input{Sections/07_conclusion}

\bibliographystyle{IEEEtran}
\bibliography{refs.bib}
\end{document}

%% file: Sections/01_intro_Lanjun.tex
\section{Introduction}
\label{sec:intro}
The recent advances in the field of Artificial Intelligence (AI) led to the creation of new applications in computer vision, natural language processing, signal processing, etc. Among these applications, image-to-image translation (image translation in short) models have gained increasing attentions. Image translation models map images from one domain to images in another domain. They are used for data synthesis (dataset augmentation), image colorization, pose transfer~\cite{pix2pix}, etc. These mentioned use cases are based on the conditional Generative Adversarial Network (cGAN)~\cite{pix2pix,pix2pixHD} which is a commonly used model in supervised image translation tasks.
%for mapping one image that is given as a condition to the model to the output image.

The models may leak information of the training dataset, even with black-box access to them~\cite{shokri2017membership}. Meanwhile, the practical use cases of the image translation models including the models trained with sensitive data (e.g., a medical dataset)~\cite{yang2020mri} continue to expand. %as practical usages of the image translation models keep increasing, the \saeed{technique has been applied on use cases that models trained with sensitive data (e.g., a medical dataset)~\cite{yang2020mri}.}  As a result, 
Hence, privacy concerns arise in relation to image translation models.

%Deep models may leak information about the datasets or the model, even with black-box access to the model. Therefore, as the practical usage of the image translation models increases, including models trained with sensitive data such as medical dataset, the privacy concerns related to these model raises.

Membership Inference Attacks (MIAs) is a way to evaluate the training data leakage (also known as membership leakage) of a model where an adversary tries to find out whether a given sample is used in training of the model or not~\cite{shokri2017membership}. Membership status is considered as personal data~\cite{law}, so it should be kept privately. However, it is recently shown in~\cite{ICCV'21} that the supervised image translation models, which are the focus of this paper, are vulnerable to MIAs, and an adversary can predict the membership status of the samples with a high success rate.

Moreover, previous studies~\cite{monte_carlo,overfitting} summarize that a major root cause of the membership leakage is overfitting.  
Especially in generative models, the training data memorization is more serious due to a small amount of training samples~\cite{gan-limited-1,gan-limited-2} and the uncertainty of the outputs~\cite{ICCV'21}. As a result, a defense method against MIAs is required to avoid overfitting and obtain a high level of output utility. %good utility of the output.  

% In addition, overfitting causes dataset memorization which is a key challenge in training generative models (including the image-to-image translation models). Therefore the reduction of overfitting helps improving the privacy of the model as well as avoiding the model to memorize data directly from the observed samples during the training.

% However,  image to image translation models are highly susceptible to leak membership status of the data samples. Authors in~\cite{ICCV'21} also show that the applying number of widely used defenses for improving the privacy of deep learning models either did not either improve the privacy of the models or generate high quality images. 

In this paper, we aim to propose a defense method against MIAs for image-to-image translation models, which achieves a better tradeoff between the output utility and membership privacy protection.  
Specifically, besides preventing the threat of MIAs, the proposed defense method obtains models with good generalizability and high quality output images.
% at the same privacy level, the proposed defense method has to generate higher quality images than those generated by other defenses.   
% In other words, at the same privacy level, the proposed defense based on Adversarial Knowledge Distillation (AKD) generates higher quality images compared to other defenses. 
%In other words, at the same privacy level, the proposed defense based on Adversarial Knowledge Distillation (AKD) generates higher quality images compared to other defenses. 

%To achieve utility-privacy tradeoff is a challenging task for existing defense approaches.  
Existing defence techniques struggle to establish a utility-privacy tradeoff for image translation models. 
% For example, \cite{ICCV'21} has shown that the applying number of widely used defenses for improving the privacy protection of deep learning models either reduce the quality of generated images significantly or are ineffective to protect the model. 
For example, it is shown in~ \cite{ICCV'21} %has shown 
that existing defense methods fail to protect the membership privacy in the image translation task. 
Besides the defense methods used in~\cite{ICCV'21}, knowledge distillation based methods are emerging techniques to mitigate MIAs in the image classification task~\cite{DMP,tang2021mitigating,zheng2021resisting}.  However, it is not trivial to apply them directly on image translation tasks. 
As the sample sizes of datasets in image translation are generally smaller than those in classification, %to apply applying the dataset splitting step and then use subsets in 
splitting the data and using subsets in
training as in~\cite{tang2021mitigating,zheng2021resisting} brings more risks of overfitting.  Moreover, the major task of the private teacher model in~\cite{DMP} is to select the data by the entropy of the prediction, but the output of an image translation task does not have such entropy information, which indicates~\cite{DMP} cannot be fully utilized on image translation.  
Therefore, we have to consider a task-specific component in designing the defense method against MIAs in image translation.

In order to achieve a better utility-privacy tradeoff, we propose a method, Adversarial Knowledge Distillation (AKD), which combines the adversarial training~\cite{GAN} with knowledge distillation mechanism~\cite{hinton2015distilling}. % to defend  MIAs on image translation. 
In details, a private model is initially trained with the private dataset. Then, the private model is used as the teacher model to train a student model in an adversarial manner with a proxy dataset. The proxy dataset is a set of unlabelled input domain images. Finally, the student model is deployed as a public model to prevent the leakage of the private dataset. To the best of our knowledge, AKD is the first defense method on image translation models that protects membership privacy while generates realistic outputs similar to the outputs from models trained without any constraint on privacy protection.
% In the proposed method, the private model is initially trained with the private dataset then used as a teacher model to distill knowledge to a student model in an adversarial manner using a proxy dataset.
% The proxy dataset is a set of unlabelled input images for which the teacher model generates the label. 
%In addition, triplet loss is used as a regularizer in the distillation process that helps to further mitigate MIA by forcing a margin between the generated samples and samples from a real dataset.

% Although knowledge distillation is used as a defense method against MIA in the earlier work, Distillation for Membership Privacy (DMP)~\cite{DMP}, the proposed knowledge-distillation based defense method in this paper is different from DMP in multiple aspects. First, the proposed method uses a adversarial distillation method including a discriminator in the training that helps the generator to create images with higher quality. The other difference is that DMP is proposed for image classification models, and the unlabelled data is sampled before distillation using the entropy of the logits at the model output. However, such approach is not applicable to the image-to-image translation models. In our method, no pre-processing is needed to be applied to the unlabelled data.

%  We show that qualitatively and quantitatively that the adversarial knowledge distillation method reduces overfitting by reducing the generalization gap of image to image translation models.

The contributions of this study are summarized as:
\begin{itemize}
  \item We propose a novel method of knowledge distillation for image translation models which is used as a defense method against MIAs.  The proposed defense method is a post-hoc method, meaning that it can be applied to models that have already been trained.
  %\item \saeed{Use of triplet loss is proposed a regularization method that reduces the membership inference attack}
  \item We show qualitatively and quantitatively that %
  adversarial knowledge distillation %the proposed method 
  %avoids 
  reduces overfitting, %in image translation models, %the regular training model.  %The method can improve privacy protection and reduce the data memorization in the image translation models.  
  and thus provides stronger protection against MIAs.
  %and reduces the data memorization in the image translation models.
 
  \item Through extensive experiments we show that our proposed method achieves the state-of-the-art tradeoff between the quality of the generated samples and the membership privacy protection for image translation models. 
\end{itemize}

The remainder of the paper is organized as follows: preliminaries are discussed in Section~\ref{sec:prem}. The proposed method is introduced in Section~\ref{sec:proposed}. The experimental setup is discussed in Section~\ref{sec:exp_setup} followed by the experiments in Section~\ref{sec:exp}. Section~\ref{sec:related_work} briefly reviews the related works, and Section~\ref{sec:conc} concludes the paper.

%todo: 
%mention LOGAN

%% file: Sections/02_preliminary.tex
\section{Preliminaries}
\label{sec:prem}
\subsection{Image-to-image Translation}
Let $S_{train} = \{(\mathbf{x}^{(n)},\mathbf{y}^{(n)})\}_{n=1}^{N}$ be {the training set of an image translation model, and $\mathcal{D}$ be the distribution that $S_{train}$ is sampled from. %a set of $N$ image pairs \wanglj
$\mathbf{x}^{(n)}$ is an image from the input domain and $\mathbf{y}^{(n)}$ is the corresponding ground truth from the output domain,}  %\wanglj{A given image translation model $G$ is trained with the goal of targeting to generate $\textbf{y}$ by the input instance $\textbf{x}$.} %, i.e., $\textbf{y}=G(\textbf{x})$.
% {The goal of training an image translation model $G$ is to get it to generate $\mathbf{y}^{(n)}$ from the input $\mathbf{x}^{(n)}$.}
% {The goal of an image translation model $G$ is to generate $\mathbf{y}^{(n)}$ from the input $\mathbf{x}^{(n)}$.}
The goal of an image translation model $G$ is to generate an image in the output domain by a given input.
The conditional Generative Adversarial Network (cGAN) is frequently used as an image translation model where the output image is conditioned on the input (~\cite{pix2pix},\cite{ pix2pixHD},\cite{img_trans_exp3},\cite{img_trans_exp4}).

%A popular choice for the mapping function is to use generative models based on Generative Adversarial Networks (GAN) in which generator $G$ is trained in the presence of an adversarially trained discriminator.  

\begin{figure}[t!] 
\centering
\includegraphics[width=\columnwidth]{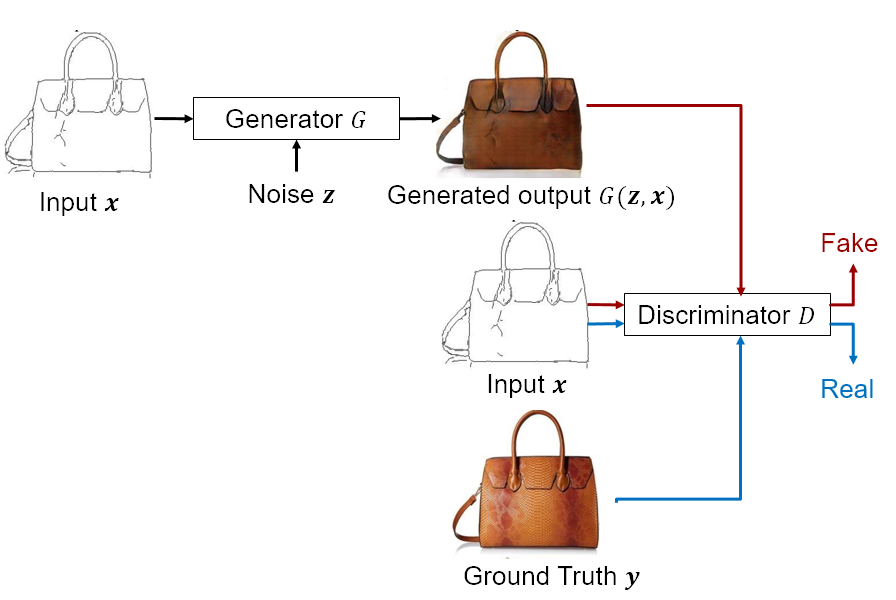}
\caption{An example to demonstrate an image translation model based on the cGAN. After training, the generator $G$ is deployed to generate images, and the discriminator $D$ is discarded. The example figure is from~\cite{pix2pix}.}
\label{fig:im2im}
\end{figure}

The conditional generative models are generally trained to map the input image $\mathbf{x}$ and a noise vector $\textbf{z}$ to the ground truth image $\textbf{y}$. An example for cGAN is shown in Fig.~\ref{fig:im2im}. The generator $G$ is trained based on the feedback from the discriminator $D$, %\wanglj{
where $D$ determines whether a given sample is real. That is to say, $D$ is trained to maximize the probability of assigning the correct real/fake label to real samples (i.e., the ground truth) and the generated samples from the generator $G$. Meanwhile, the generator $G$ is simultaneously trained to generate samples to fool $D$.
%} 
The loss function to train a cGAN can be defined as:
%\imp{
 \begin{equation}
 \begin{aligned}\label{eq:cgan_loss}
     L_{cGAN}(G,D)= & \mathbb{E}_{\mathbf{y,x}}[ \log(D(\mathbf{y}, \mathbf{x})) ] +\\
     & \mathbb{E}_{\mathbf{z,x}} [\log(1 - D(G(\mathbf{z},\mathbf{x}),\mathbf{x}))]
\end{aligned}
 \end{equation}
The objective is to obtain an optimal generator $G^*$ as:
\begin{equation}
    G^* = \operatorname*{arg} \min_{G} \max_{D} L_{cGAN}(G,D)
\end{equation}
%}
where $D$ tries to maximize Eq.\eqref{eq:cgan_loss} while $G$ aims at minimizing it. 
% Therefore, the problem is a min-max game with final objective $G^* = \operatorname*{arg} \min_{G} \max_{D} L_{GAN}(G,D)$. 
%Different generative models based on GAN are introduced that follow the same training procedure as GANs but their objectives are modified. For instance, in cGAN which is frequently used as image translation models, the output image is conditioned on the input, and both $G$ and $D$ receive the input $\mathbf{x}$ as a condition. \imp{For further details on GAN and cGAN please refer to~\cite{GAN,pix2pix}.}

%\saeed{Needs a little bit details about GAN}
%\subsection{Threat model}
%\saeed{In some papers they have a section called threat model which is similar to my next subsection. We might need to adjust it.}

\subsection{MIA on Image Translation Models}
\label{subsec:MIA}
In the MIA, an adversary's goal is to find out whether a given sample ($\mathbf{x}$,$\mathbf{y}$) is used in the training of the target model. 
\begin{definition}[Membership Inference Attack]
\label{def:MIA}
Given a target model $G$, %the external knowledge of adversary $\mathcal{K}$
and a data point $(\mathbf{x},\mathbf{y})$, the membership inference attack is defined as an adversary determining whether $(\mathbf{x},\mathbf{y}) \in S_{train}$ or $(\mathbf{x},\mathbf{y}) \notin S_{train}$.  
\end{definition}

The MIA is performed either in black-box or white-box settings~\cite{nasr_S&P}. In the black-box setting, an adversary can only access the model output. Specifically, given the generator $G$, the adversary decides about the membership status of the given sample by inspecting the model output $G(\mathbf{z}, \mathbf{x})$. On the other hand, in white-box MIAs, an adversary has access to the model's weights and the intermediate features. Since the black-box attack is more practical given that the model architecture and weights are not always available, our focus is on black-box MIAs.       

Reconstruction loss-based attack is shown to achieve high MIA performance on image translation models~\cite{ICCV'21}. For a given query sample $(\mathbf{x},\mathbf{y})$ and the target generator $G$, the reconstruction loss is defined as the $\ell_1$ norm of the difference between the model output $G(\mathbf{z}, \mathbf{x})$ and the ground truth $\mathbf{y}$:
\begin{equation}
\label{eq:recon_loss}
    L_{rec}(\mathbf{x},\mathbf{y}) = \norm{G(\mathbf{z}, \mathbf{x})-\mathbf{y}}_{1}
\end{equation}
%\wanglj{
To determine whether the query sample $(\mathbf{x},\mathbf{y})$ belongs to the training dataset of $G$, the reconstruction loss-based attack relies on a pre-defined threshold  $\tau$.  In details, if the sample satisfies $L_{rec}(\mathbf{x},\mathbf{y})<\tau$, then it is considered as a member; otherwise, a non-member.
%}
% $L_{rec}$ is computed for all the query samples. The computed loss values are compared against a pre-defined threshold $\tau$. Samples that satisfy $L_{rec}(\mathbf{x},\mathbf{y})<\tau$ are considered as members and the rest are considered as non-members. 

The intuition behind the reconstruction loss-based attack is that %the reconstruction loss of members are generally smaller than those of non-member samples.
member samples generally have smaller reconstruction loss compared to non-member samples.
%training data generally have smaller reconstruction errors compared to that
%of non-member data.
%enerated images by the generator are generally closer to the training samples compared with the non-training samples.
The difference between the reconstruction loss on member samples and non-member samples is more apparent on overfitting models where the gap between the performance of the model on the training samples and non-training samples is larger. Recent works show that generative models {trained on small datasets} suffer from overfitting~\cite{gan-limited-1,gan-limited-2}, which makes them vulnerable to MIAs.

%% file: Sections/03_proposed.tex
\section{Proposed method}
\label{sec:proposed}
\begin{figure}[t!] 
\centering
\includegraphics[width=\columnwidth]{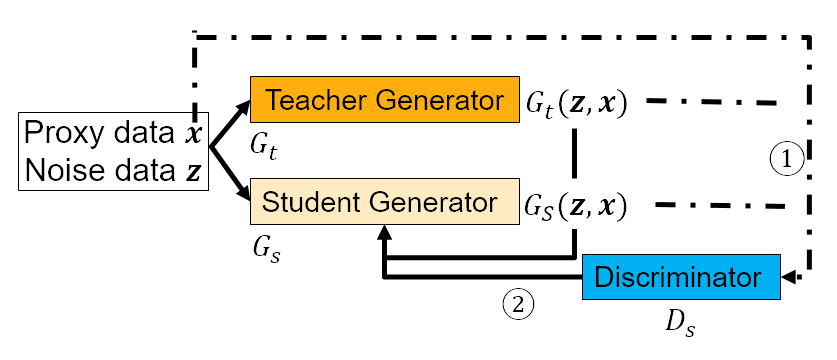}
\caption{The proposed AKD framework. %With an input $\mathbf{x}$ as well as a noise $\mathbf{z}$,
Given $\mathbf{x}$ and the noise vector $\mathbf{z}$, \numcircledmod{1} the teacher generated sample $G_t(\mathbf{z,x})$ and the student generated samples $G_s(\mathbf{z,x})$ are used to train $D_s$ by Eq.(\ref{eq:disc_loss}). Then, \numcircledmod{2} $D_s$'s output and the difference between $G_t(\mathbf{z,x})$ and  $G_s(\mathbf{z,x})$  are used to train $G_s$ by Eq.\eqref{eq:generator_loss}.} 
\label{fig:AKD} 
\end{figure}

Adversarial Knowledge Distillation(AKD) for image translation models (shown in Fig.~\ref{fig:AKD}) consists of three components: a given teacher generator $G_t$ which is to be protected, a student generator $G_s$ which is to be publicly deployed and a student discriminator $D_s$ used in the adversarial training of $G_s$. %\wanglj{adversarially}. 

The teacher generator $G_t$ is from the cGAN method which is trained on the private training dataset without any constraint on the privacy protection. 
As mentioned in Sec.~\ref{sec:prem},  in the training stage, the cGAN trains a generator and a discriminator simultaneously, while in the deployment stage, the discriminator is generally discarded and only the generator is used to generate images. 
% The teacher model's  after training, and only the generator is deployed to generate images in practice. 
% As mentioned in Sec.~\ref{sec:prem}, to train a cGAN requires a generator and a discriminator. The teacher model's discriminator is generally discarded after training, and only the generator is deployed to generate images in practice. 
Therefore, it is practical to suppose that the discriminator of the teacher model (i.e. the model to be protected) does not exist even for the MIA defense purpose, and the generator ($G_t$) is only available for queries (i.e. the model architecture and weights are not necessary). 
% Hence, the proposed method can be applied to any given black-box image translation generator.  
In this study, we design to use this teacher generator to generate labels for unlabelled samples from the proxy dataset in order to train the student generator and the student discriminator.  

The proxy dataset $S_{proxy} = \{(\mathbf{x}_p^{(n)})\}_{n=1}^{P}$ is a set of unlabelled data samples. The samples in the proxy dataset can be taken from a publicly available dataset.
% In case there is no publicly available dataset, it is possible to create it by a generative model trained using the samples in the private dataset. 
%For instance,  %StyleGAN2~\cite{proxy_gen} %is used to generate
%images that do not exist in the  training dataset of the generative model. Such generative models can be used to generate samples for the proxy dataset.
 %real-looking images of faces, objects, buildings and animals 
In case there is no publicly available dataset, it is possible to create it by a task-specific generative model %\wanglj{related to
trained to generate the samples corresponding to the input domain of the private dataset (i.e., without any information related to the output domain).
For example StyleGAN\cite{stylegan} is used in~\cite{stylegan-example} to generate chest X-rays. 
% Such model can be used to generate the samples for the proxy dataset. 

Note that the above task-specific generative model (e.g. StyleGAN),  is only used to create samples for the proxy dataset, but not deployed publicly. Therefore, the privacy of that generative model is not a concern. 

Also note that an adversary might have access to the proxy samples because of public availability, but he does not have access to the private teacher model $G_t$. Hence, the output domain label corresponding to a proxy sample cannot be obtained, and the proxy dataset poses no membership privacy threat. In addition, we verify no privacy leakage on the proxy dataset by experiments as shown in Sec.\ref{subsec:exp_prox}. 

%\imp{or can be obtained by applying a training a generative model using the available samples in the private training dataset.}  %Since the labels corresponding to the samples in $S_{train}$ are not used in training the student generator, privacy of such pre-trained model is not a concern.} In our experiments, we used validation set (excluding the labels) as a proxy dataset. 

Although the student model is generally lighter than the teacher model in a distillation framework %for the compression purpose
~\cite{hinton2015distilling}, the student generator does not have such constraint. This is because we use distillation as a defense method rather than a method for model compression. Hence, there is no constraint on the student generator's capacity.  
% In the experiment, we used the same architecture for the teacher  and student generator. 

Th student generator $G_s$ and the student discriminator $D_s$ are trained iteratively which is similar to the training  procedure of the conditional generative adversarial networks as introduced in Sec.~\ref{sec:prem}. In the following, we will discuss the loss function of the student discriminator and the student generator, respectively.

\textbf{Student Discriminator} The student discriminator's goal is to distinguish the samples generated by $G_t$ and $G_s$. 
% On the other hand, $G_s$ tries fooling $D_s$ by generating samples that are close enough to the corresponding teacher outputs. 
The discriminator is a binary classification model similar as~\cite{pix2pix} by replacing the ground truth $\mathbf{y}$ with the teacher generated sample. The loss function to train the discriminator is:  
% \begin{equation}
% \label{eq:disc_loss}
% \begin{split}
%     L_{D_s}(x_p^{(n)}) = \frac{1}{2} (BCE(D_s (\overline{\overline{y}}^{(n)}_p),False) \\
%     + BCE(D_s (\overline{y}^{(n)}_p),True)) 
% \end{split}
% \end{equation}
% Where $\overline{\overline{y}}^{(n)}_p=G_{s}(x_p^{(n)})$ and $\overline{y}^{(n)}_p=G_{t}(x_p^{(n)})$ are the output of model student generator and teacher generator for input $x_p^{(n)}$, respectively. $BCE(.)$ is binary cross entropy loss, and False/True are labels.  
\begin{equation}
\label{eq:disc_loss}
\begin{aligned}
         L_{D_s}=  & \mathbb{E}_{\mathbf{z},\textbf{x}}[ \log(D_{s}(G_{t}(\mathbf{z}, \mathbf{x}), \mathbf{x})) ] +\\
     &  \mathbb{E}_{\mathbf{z},\textbf{x}} [\log(1 - D_{s}(G_{s}(\mathbf{z}, \mathbf{x}),\mathbf{x})]
\end{aligned}
\end{equation}
where $G_{s}(\mathbf{z}, \mathbf{x})$ and $G_{t}(\mathbf{z}, \mathbf{x})$ are outputs of the student generator and the teacher generator for the input $\mathbf{x}$ and a noise vector $\mathbf{z}$, respectively. %$BCE(.)$ is the binary cross entropy loss, and False/True are labels.  %\imp{ $\mathbf{z}$ is the noise in the form of dropout that is applied to several layers of the network during the training and testing.  }

\textbf{Student Generator} Student discriminator's weights are updated according to loss in Eq.\eqref{eq:disc_loss}. Then the updated discriminator is used to tune the weights of the student generator. The generator tries to generate samples that fool the discriminator, which are close enough to the teacher generated samples. Hence, the loss function for training the student generator is defined as:
\begin{equation}
\label{eq:generator_loss} 
\begin{aligned}
    L_{G_s} = & \mathbb{E}_{\textbf{z,x}}[ \log(D_{s}(G_{s}(\mathbf{z}, \mathbf{x}), \mathbf{x})) ] +\\
    &  \lambda~ \mathbb{E}_{\textbf{z,x}} [\norm{G_{s}(\mathbf{z} , \mathbf{x})-G_{t}(\mathbf{z} , \mathbf{x})}_{1}]
\end{aligned}
\end{equation}
%    L_{G_s}(\mathbf{x}_p^{(n)}) = BCE(D_s (G_{s}(\mathbf{z} , \mathbf{x}_p^{(n)}),True) \\
%    + \lambda \norm{G_{s}(\mathbf{z} , \mathbf{x}_p^{(n)})-G_{t}(\mathbf{z} , \mathbf{x}_p^{(n)})}_{1}
where $\lambda$ is a parameter for adjusting the weight of the $\ell_1$ loss between the teacher generated sample and the student generated sample. After the student generator weights are updated according to Eq.\eqref{eq:generator_loss}, the updated student generator is used to tune the weights of the discriminator. This process is repeated for each batch during the training.    

%% file: Sections/04_exp_setup.tex
\section{Experimental Setup}
\label{sec:exp_setup}
\subsection{Data Preparation for Evaluation}
The proposed defense method is evaluated against the MIA discussed in Section~\ref{subsec:MIA} using samples in attack evaluation dataset. Attack evaluation dataset $S_{attack}=S_M\cup S_{NM}$ consists of two subsets: the members set $S_M$ and the non-members set $S_{NM}$. The members set includes samples from training set, i.e., $S_M \subseteq S_{train}$, and the non-members set $S_{NM}$ is sampled from $\mathcal{D}$ that are not in $S_{train}$.  Following the earlier works in the literature~(\cite{ICCV'21}, \cite{monte_carlo}),  non-member samples are taken from testing dataset ($S_{test})$, and equal number of samples are included in $S_M$ and $S_{NM}$. It is worth mentioning that splitting the training data into smaller datasets and using the smaller dataset to train image translation models is not a practical evaluation scenario because it significantly increases the overfitting. Therefore, in the experiments, all the training samples are used to train the image translation models.

%$N'$ and $M'$ differ among the works in the literature. In majority of the works such as ~\cite{MIA_shokri, ml-leaks, RL-MIA} $N'$ is set to be equal to $M'$ in order to maximize the uncertainty of membership inference. But non-equal values are also used in some works (e.g. $N'= 5M'$ in~\cite{difficulty} and $M'= 10 N'$ in~\cite{realistic}). In the experiments we set $N'=M'$. 

\subsection{Model and Datasets}
\label{subsec:model_data}
The proposed method and the baselines are evaluated using the pix2pix image translation model~\cite{pix2pix}  on two tasks. The first task is to convert the architectural maps to real photos of buildings, and the second task is to generate real scenes from semantic maps. 

The model for the first task is trained using CMP Facade dataset~\cite{facade}. The validation set in CMP Facade dataset is used as the proxy dataset in the proposed method.  

In the second task, Cityscapes dataset~\cite{Cityscapes} is used to train the models. Since the semantic labels of testing dataset are withheld by the data provider, we do not have testing samples to check the performance of the models. %testing dataset could not be used for testing. 
Therefore, we split the validation set into two sets of the same size, and use the first set as proxy dataset and the remaining samples as test set.

The number of samples in the training dataset, proxy dataset and the testing dataset are summarized in Table~\ref{tbl:dataset}. All the samples in the testing dataset are included in $S_{NM}$, and the samples in $S_M$ are randomly selected from the training dataset.

\begin{table}
\centering
\caption{The number of samples in the datasets}
\label{tbl:dataset}
\begin{tabular}{|c|c|c|c|} 
\hline
\multicolumn{1}{|l|}{} & \multicolumn{1}{l|}{Training dataset} & \multicolumn{1}{l|}{Proxy dataset~} & \multicolumn{1}{l|}{Testing dataset}  \\ 
\hline
Facade                 & 400                                   & 100                                 & 106                                   \\ 
\hline
Cityscapes             & 2,975                                  & 250                                 & 250                                   \\
\hline
\end{tabular}
\end{table}

The student generator is a Unet-based model, and  the student discriminator is a four-layer CNN taken from the generator and the discriminator models in pix2pix model~\cite{pix2pix}. The knowledge distillation is performed using the proxy datasets for 200 epochs (same number of epochs used for training the teacher model). Following~\cite{pix2pix}, $\lambda$ in Eq.\eqref{eq:generator_loss} is set to $100$ and the batch size is set to 1 in the experiments. Also following~\cite{pix2pix}, the noise vector $\mathbf{z}$ is generated using dropout layers during the training and testing the model.
%Since the number of samples in the proxy dataset is (upto 12 times) smaller compared to the training samples, the student model training need shorter training time compared to training the teacher model. 
% In all the experiments the batch size is set to 1. 

\subsection{Baselines}
The following three defense methods are used as baselines to evaluate  %compare 
the performance of the proposed defense. %against.

\textbf{1. Gauss defense:} In Gauss defense~\cite{gauss}, the output of the generator is altered by a additive Gaussian noise. The additive noise is used to reduce the similarity between the generated samples and the ground truth. 

\textbf{2. Differential private SGD (DP-SGD) defense:} In DP-SGD~\cite{DP-SGD}, the gradients of the model parameters are clipped, then a Gaussian noise ($\mathcal{N} (0,\sigma^2)$) is added to the clipped gradients during the training to achieve a differentially private model. The variance of the added noise $(\sigma^2)$ adjusts the utility-privacy tradeoff. %Earlier works~\cite{ICCV'21,logan} reported that training a generative model in deferentially private manner causes issues such as unstable training, visually not satysfing results and longer convergence. 
%In our experiments, by running multiple combinations of clipping range and noise variance we obtained image translation models at different utility-privacy tradeoffs. We include a high utility-low privacy and low utility-high privacy cases obtained using DP-SGD method. 

\textbf{3. DMP based defense:} We devise a variant of DMP~\cite{DMP}, which is originally proposed for image classification models, to adapt the image translation task.
% Following DMP, no student discriminator is used in the modified defense. 
The difference between this baseline and the original DMP is that original DMP performs a pre-selection of samples in the proxy dataset using the entropy of the teacher model output before knowledge distillation. This is not applicable to the image translation models, because the calculation of entropy of the large dimensional output images is not trivial. Therefore, the proposed pre-processing step in DMP is not included in the the devised baseline.   
% a knowledge distillation based membership defense similar to

\subsection{Evaluation metrics}
\subsubsection{Utility-Privacy tradeoff}
A membership defense method is evaluated by two aspects. The first aspect is to measure the membership inference attack performance. Following previous studies as~\cite{pyrgelis2018knock, ICCV'21}, we use the area under the receiver operating characteristic curve (AUCROC).  Suppose a model without privacy leakage is given,  according to  %then as 
our attack evaluation dataset setting, the minimum attack AUCROC is 50\% which is obtained by random guessing.  That is to say, the closer AUCROC is to 50\%, the better privacy protection performs. 

The second aspect in evaluating a defense method is to measure the utility of the model outputs.  In the image translation task, the utility is represented by the quality of the generated samples.  
The quality of the generated samples is measured using Kernel Inception Distance (KID)~\cite{KID} which is a better metric compared to Fréchet Inception Distance (FID)~\cite{FID} when testing data is not a large scale set~\cite{gan-limited-1}. Note a lower KID is the indicator of a higher quality. %\wanglj{In addition, it is noted that the lower KID, the better performance on the image quality.}

%\wanglj{However, to evaluate}
Although the utility and the privacy protection performance can be evaluated for a membership defense method, calculating the utility-privacy tradeoff for the image translation task is not straight forward due to the fact that metrics (AUCROC and KID) have different scales, whereas in classification models both the utility (e.g., model top-1 prediction accuracy) and the privacy protection performance (i.e., AUCROC) are on the same scale. 
Thus, to measure the utility-privacy tradeoff for image translation tasks,  we design the Normalized KID (NKID) to bring it within $0\%\leq \text{NKID} \leq 100\%$. Since the KID is a distance type metric for which lower means a better quality, we normalize KID by setting the worst quality for the generated images ($\text{KID}_{max}$) to 100\%. Then the KID is scaled to obtain the NKID as $\text{NKID} = \frac{\text{KID}}{\text{KID}_{max}}$ for each experiment. 
% We replaced model output by random Gaussian noise at different variance and constant values to find the largest KID. 

After %testing various approaches, such as 
measuring KIDs %s by 
for images with random Gaussian noises under different variances as well as images with constant values, %etc., 
we find black images resulted in the largest KID, i.e. $\text{KID}_{max}$, and $\text{NKID}$ is calculated based on it. %}
% Among them, generating . 
To sum up, by normalizing the KID, AUCROC and NKID are on the same scale and the utility-privacy tradeoff can be compared among different defense methods. 

Model outputs slightly change from one training to another due to random initialization. Hence, the membership inference attack performance slightly varies among the models trained with different initialization. We repeat training and testing for all the experiments five times and report the average performance for the proposed method and the baselines. 

\subsubsection{Generalization Gap}
The generalization gap is closely associated with overfitting to the training dataset. It is defined as the difference between a model's performance on its training data and its performance on the unseen data from the same distribution (e,g. testing dataset)~\cite{par-gan}. The generalization gap $g$ for an image translation model $G$ is measured as:
\begin{equation}
\label{eq:gen_gap}
g(G) = Q_{\mathbf{x} \in {X_{train}}} (G(\mathbf{z}, \mathbf{x})) - Q_{\mathbf{x} \in {X_{test}}} (G(\mathbf{z}, \mathbf{x}))
\end{equation}
where $Q(.)$ measures the quality of the generated samples which %we use  
is NKID in this study. $X_{train}=\{\mathbf{x}^{(n)}\}_{n=1}^{N}$ is the input domain samples in $S_{train}$ and $X_{test}=\{\mathbf{x}^{(n)}\}_{n=1}^{M}$ is the input domain samples  %from 
in $S_{test}$ where $M$ is the number of samples in the testing dataset. We show in the next section how the generalization gap changes after applying the proposed defense.
% \begin{equation}
% \label{eq:gen_gap}
% g(G) = \frac{1}{N}\sum_{\mathbf{x} \in {X_{train}}} \text{NKID}(G(\mathbf{z}, \mathbf{x})) - \frac{1}{M}\sum_{\mathbf{x} \in {X_{test}}} \text{NKID}(G(\mathbf{z}, \mathbf{x}))
% \end{equation}

%% file: Sections/05_exp.tex
\section{Experiments}
\label{sec:exp}
In this section, we first provide the experiments related to the utility-privacy tradeoff for the proposed method and the baselines. As mentioned in Section~\ref{sec:exp_setup}, the experiments are conducted using pix2pix~\cite{pix2pix} on two tasks. 
We also confirm that there is no privacy leakage on the proxy dataset we used to train student generators.
%\wanglj{We also verify no privacy leakage on the proxy dataset we used in training student generators.}
Furthermore, the generalization gap of the proposed method is discussed in comparison with the regular training model.   
\subsection{Utility-privacy tradeoff}
\subsubsection{Regular training}
\label{subsubsec:reg_exp}
In the first experiment, the performance of the attack on the model trained with no defense is analyzed. 
% Training with the training dataset without performing defence during the training or after training is the default training setting for the original image translation models in~\cite{pix2pix}. 
Regular training refers to a training method without conducting any defence during or after the training.
%\wanglj{The regular training is to train the model without performing defence during the training or after training on image translation models in~\cite{pix2pix}. }
The average NKID and AUCROC of two tasks for the regular training are reported in Table~\ref{tbl:result_facade} and Table~\ref{tbl:result_city}. As it can be seen in the tables, the attack reaches a high AUCROC on the model obtained by regular training. This verifies the findings in~\cite{ICCV'21} about the vulnerability of the image translation models to MIAs. 

% \begin{table}[t!]
% \centering
% \caption{Mean and standard deviation of the attack accuracy and the quality of the generated images for the tested models on the first task (architectural maps $\rightarrow$ real photos of building) trained using CMP Facade dataset}
% \label{tbl:result_facade}
% \begin{tabular}{|c|c|c|c|}
% \hline
% \textbf{}        &AUCROC   & NKID               \\ \hline
% Regular   Training   &96.46 $\pm$ 0.44 & 5.93  $\pm$ 0.22          \\\hline
% Gauss Defence           & 96.40 $\pm$ 0.47 & 6.49  $\pm$ 0.31           \\ \hline
% DP-SGD~($\sigma=2.02\times10^{-3}$)      & 89.20 $\pm$ 0.69 & 8.21  $\pm$ 0.39            \\\hline
% DP-SGD~($\sigma=1.69\times10^{-2}$)       & 58.06 $\pm$ 0.94 & 12.57 $\pm$ 0.46         \\\hline
% DMP                 & 57.73 $\pm$ 0.63 & 13.64 $\pm$ 0.68          \\\hline
% AKD                 & 57.56 $\pm$ 0.40 & 7.80  $\pm$ 0.33       \\\hline
% %P =1, mar =   0.00 & 58.36 $\pm$ 0.24 & 8.13  $\pm$  0.17         \\\hline
% %P =1, mar =   0.05 & 57.97 $\pm$ 1.01 & 7.84  $\pm$ 0.41      \\\hline

% \end{tabular}
% \end{table}

\begin{table}[t!]
\centering
\caption{Mean and standard deviation of the attack performance and the quality of the generated images for the tested models on the first task (architectural maps $\rightarrow$ real photos of building) trained using CMP Facade dataset}
\label{tbl:result_facade}
\begin{tabular}{c|c c}

\textbf{}        &AUCROC   & NKID               \\ \hline
Regular   Training   &96.46 $\pm$ 0.44 & 5.93  $\pm$ 0.22          \\
Gauss Defence           & 96.40 $\pm$ 0.47 & 6.49  $\pm$ 0.31           \\ 
DP-SGD~($\sigma=2.02\times10^{-3}$)      & 89.20 $\pm$ 0.69 & 8.21  $\pm$ 0.39            \\
DP-SGD~($\sigma=1.69\times10^{-2}$)       & 58.06 $\pm$ 0.94 & 12.57 $\pm$ 0.46         \\
DMP                 & 57.73 $\pm$ 0.63 & 13.64 $\pm$ 0.68          \\\hline
AKD                 & 57.56 $\pm$ 0.40 & 7.80  $\pm$ 0.33       \\
%P =1, mar =   0.00 & 58.36 $\pm$ 0.24 & 8.13  $\pm$  0.17         \\\hline
%P =1, mar =   0.05 & 57.97 $\pm$ 1.01 & 7.84  $\pm$ 0.41      \\\hline

\end{tabular}
\end{table}

\begin{table}[t!]
\centering
\caption{Mean and standard deviation of the attack performance and the quality of the generated images for the tested models on the second task (semantic maps $\rightarrow$ real photos of outdoor scenes) trained using Cityscapes dataset}
\label{tbl:result_city}
\begin{tabular}{c|cc}
\textbf{}               & AUCROC & NKID    \\ \hline
Regular   Training      & 87.19 $\pm$ 0.35  & 3.26 $\pm$ 0.23     \\ 
Gauss Defence               & 87.22 $\pm$ 0.62  & 3.88  $\pm$ 0.20     \\ 
DP-SGD~($\sigma=1.0\times10^{-3}$)            & 86.07 $\pm$ 0.56 & 4.90  $\pm$ 0.68     \\ 
DP-SGD~($\sigma=1.85\times10^{-2}$)           & 62.80 $\pm$ 1.55 & 27.57 $\pm$ 4.21     \\ 
DMP                     & 62.56 $\pm$ 0.70 & 12.80 $\pm$ 0.48     \\ \hline
AKD                     & 62.35 $\pm$ 0.56 & 4.66  $\pm$ 0.30     \\ 
%P =1, mar =   0.00      & 62.20 $\pm$ 0.53 & 4.64  $\pm$ 0.40    \\ \hline
%P =1, mar =   0.05      & 62.72 $\pm$ 0.74 & 4.82  $\pm$ 1.11     \\ \hline
\end{tabular}
\end{table}

\subsubsection{Gauss defense} In Gauss defense, additive zero mean Gaussian noise is added to the model output. The AUCROC and NKID with respect to the noise variance are plotted in  Fig.~\ref{fig:facade_points} and Fig.~\ref{fig:city_points}. Note that the coordinate origin (i.e. AUCROC is 50\% and NKID is 0\%) represents the ideal optimal utility-privacy tradeoff, and the point more closer to the origin indicates a better trade-off performance.  As illustrated in the figures, %the AUCROC is not reducing with respect to noise variance values increasing. 
for Gauss defense, the AUCROC does not decrease as the noise variances increase. 
Hence, to defend MIAs by Gauss defense is not feasible. {%We also include 
% Hence, the MIA performance is not reduced using the Gauss defense, and defending MIAs using Gauss defense is not feasible. {%We also include 
In addition, we also include Gauss defenses in Table~\ref{tbl:result_facade} and Table~\ref{tbl:result_city} for comparing the results against other baselines.}  %that Figures also show that applying Gaussian defense degrades the quality of the generated images.

% \begin{figure}[t!] 
% \centering
% \includegraphics[width=0.7\columnwidth]{./figures/gauss-def}
% \caption{Applying Gaussian Defense against MIA on two tasks }
% \label{fig:gauss-def}
% \end{figure}

\subsubsection{DP-SGD} 
We train a set of models with DP-SGD by changing the noise variance. The utility-privacy of the trained models by DP-SGD are shown in Fig.~\ref{fig:facade_points} and Fig.~\ref{fig:city_points} for two tasks. As the variance of the added noise to gradient increases, the AUCROC increases and the utility decreases. 

Among the models with DP-SGD, we pick two models for the purpose of comparing against AKD. The  AUCROC, NKID and the corresponding standard deviation of the noise in the chosen models are shown in Table~\ref{tbl:result_facade} and Table~\ref{tbl:result_city}.  
The first model is chosen because its utility level is close to the utility of AKD. Hence, the privacy protection performance of DP-SGD and AKD can be compared at a comparable utility level. %  (roughly) the same 

The second model is chosen because it achieves the privacy protection performance close to that obtained by AKD. Therefore, the quality of the generated samples using DP-SGD and AKD can be compared at similar privacy level.

%For both models, clipping threshold is set to 1 but higher noise variance is chosen for low utility-high privacy model. The variance of the noise is chosen such that the utility and privacy of the models obtained from DP-SGD roughly matches our proposed model's performance so that they can be compared at the same utility and privacy levels. 

%Comparing the two models obtain from DP-SGD in each task reveals that adding noise to gradients with larger variance improves the privacy at the cost of degrading the quality of the generated samples. In the first task (Table~\ref{tbl:result_facade}), the quality degradation is minor when membership privacy is increased. However, in the second task the quality drops significantly when the membership privacy is improved by increasing the noise variance. 

\subsubsection{DMP-based} 
In DMP, the private teacher model from the regular training (Section~\ref{subsubsec:reg_exp}) is used to distill knowledge to the student model without using a discriminator in the distillation process. As shown in Table~\ref{tbl:result_facade} and Table~\ref{tbl:result_city}, the DMP-based method reduces the AUCROC while increases the NKID compared to the regular training. 

Next, we discuss the performance of the proposed method compared to the baselines including the DMP-based defense and we show that the proposed method generates images at higher quality compared with DMP-based method.

\subsubsection{Adversarial Knowledge Distillation (AKD)} 
The teacher model used in the proposed AKD is the regular model discussed in Section~\ref{subsubsec:reg_exp}. %For the samples in the unlabelled proxy dataset, the teacher model generates pseudo labels which are used to distill knowledge from teacher model to the student model in the presence of a discriminator. 
The AUCROC and NKID %the measured generated images' quality 
for the student model trained for the first and second tasks are shown in Table~\ref{tbl:result_facade} and Table~\ref{tbl:result_city}, respectively. 

The tables indicate that the AKD significantly reduces the AUCROC (up to 38.89\%) against the MIA without a noticeable increase in the NKID compared with the regular training. Note that the MIA reaches high attack performance on the teacher model as discussed in Section~\ref{subsubsec:reg_exp}. Nonetheless, the AKD mitigates the attack performance on the student model. 

\begin{figure}[t!] 
\centering\includegraphics[width=0.9\columnwidth]{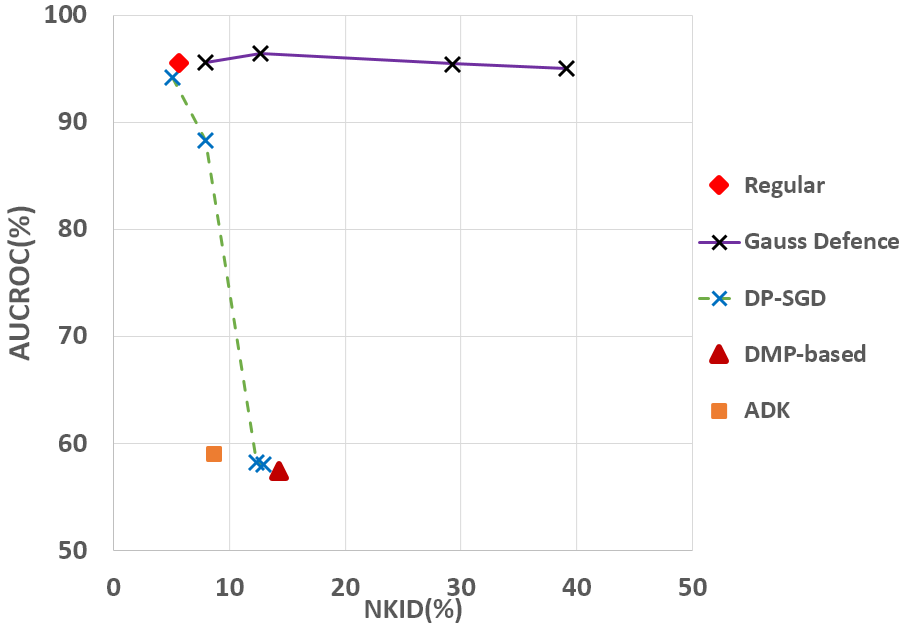}
\caption{Utility-privacy tradeoff  on Facade}
\label{fig:facade_points}
\end{figure}

\begin{figure}[t!] 
\centering
\includegraphics[width=0.9\columnwidth]{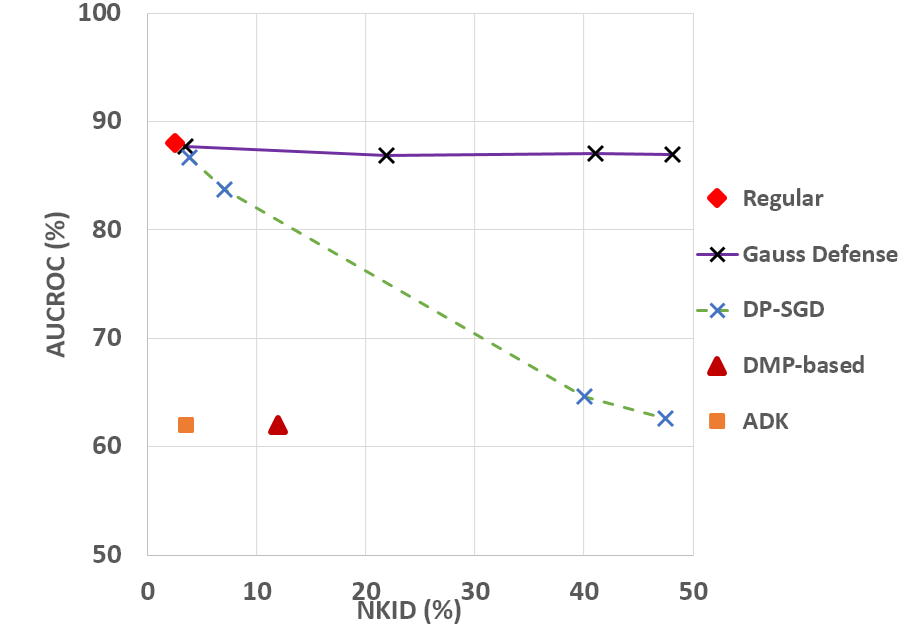}
\caption{Utility-privacy tradeoff on Cityscapes }
\label{fig:city_points}
\end{figure}

Table~\ref{tbl:result_facade} and Table~\ref{tbl:result_city} as well as the results in Fig.~\ref{fig:facade_points} and Fig.~\ref{fig:city_points} also reveal that AKD achieves the lowest AUCROC among methods with similar image qualities. %utility levels. 
Take Table~\ref{tbl:result_facade} as an example, the NKIDs of AKD, DP-SGD under $\sigma=2.02\times10^{-3}$ as well as Gauss defense are similar, but the AUCROC of the AKD is over 30\% lower than those two methods, which implies a high level of privacy protection against MIAs.
%strong performance on defense MIAs.}

Comparing  AKD with the baselines at the same privacy level shows that  AKD achieves the best quality of the generated samples. This illustrates that AKD achieves the best utility-privacy tradeoff among the tested methods. For example, in Tabel~~\ref{tbl:result_city}, by comparing the AKD with DP-SGD under $\sigma=1.85\times10^{-2}$ and DMP, all these three method achieve AUCROC around 62\%, but the NKID of the AKD is 63\% lower %than
compared to the best performing baseline.

\begin{figure*}[t!]
\centering
\includegraphics[width=1.5\columnwidth]{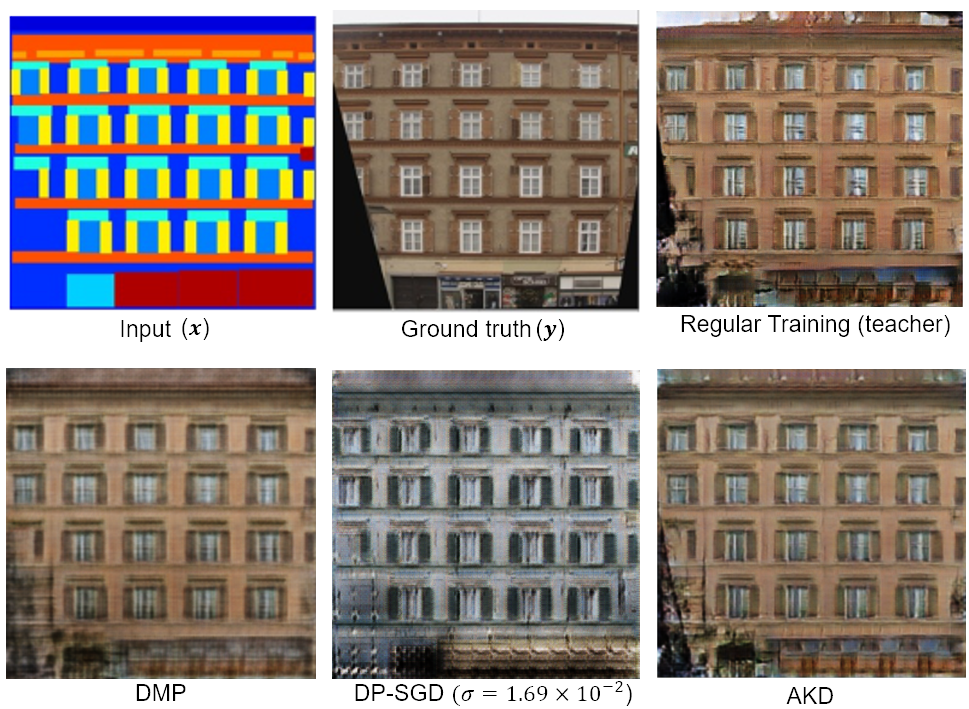}
\caption{An example from Facade}
\label{fig:visual_exp1}
\end{figure*}

\begin{figure*}[t!]
\centering
\includegraphics[width=1.5\columnwidth]{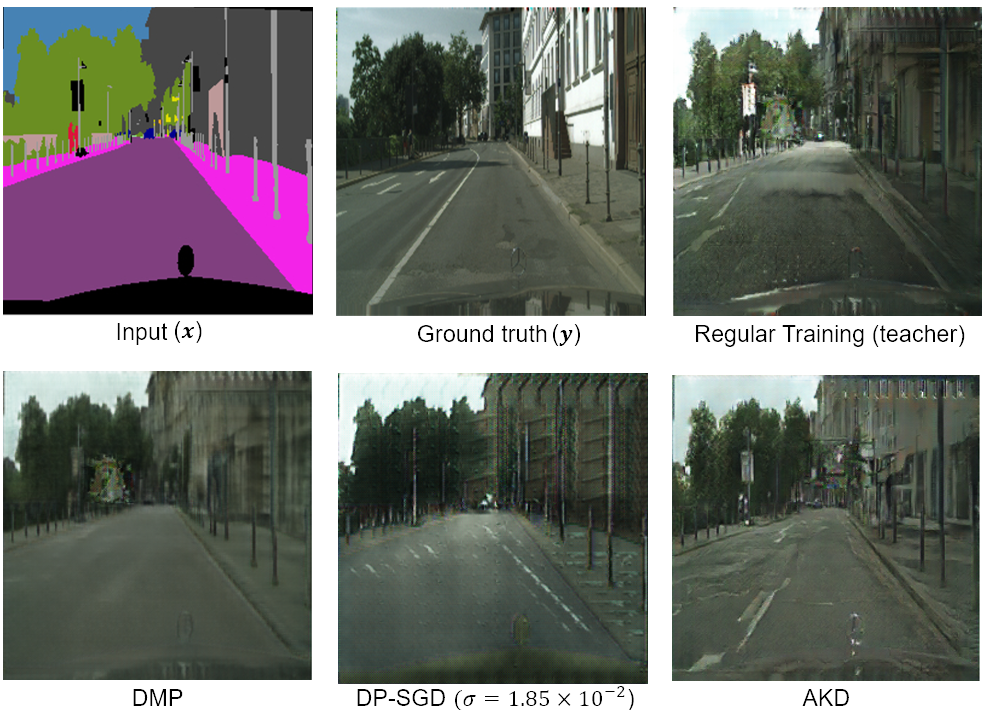}
\caption{An example from Cityscapes }
\label{fig:visual_exp2}
\end{figure*}

The case studies of the generated images for two tasks using AKD, DP-SGD and DMP-based methods are shown in Fig.~\ref{fig:visual_exp1} and Fig.~\ref{fig:visual_exp2}. The generated images are taken from the models at roughly the same membership privacy protection level. 

As illustrated in Fig.~\ref{fig:visual_exp1}, the image generated by the model obtained using the DMP-based defense is blurry and lacks details. In the image generated by the DP-SGD method, the windows are disappeared in some regions and it seems network struggles filling the details.  However, the generated image from AKD does not have the mentioned problems, and it is the most similar image to the teacher generated image. The similar pattern is also observed in the generated images in Fig.~\ref{fig:visual_exp2}, which indicates the student model obtained by ADK generates images that are visually similar to the private teacher model while reducing the model's membership leakage.


\subsection{Membership leakage on proxy dataset} \label{subsec:exp_prox}
%As the public student generator is trained by the proxy dataset, there is a potential risk on the membership leakage of the proxy dataset. 

Since the proxy dataset is used to train the publicly deployed student generator, there is a possibility of the proxy dataset's membership leakage.
Although the proxy dataset is public or self-created, it is better to avoid the {membership} leakage to protect the privacy of the corresponding data. 

In order to %check 
evaluate the leakage of the proxy dataset, we perform the MIA on student generator by considering the proxy dataset (including original labels) as member samples. The MIA AUCROC for Facade and Cityscapes are 50.63$\%$ and 51.04$\%$, respectively, which are close to the AUCROC of random guessing as we expected. As a result, the risk of proxy data membership leakage by the student generator can be ignored.  

%\subsubsection{AKD + Triplet loss}
%\saeed{I am skipping this}

\subsection{Improving the generalization gap}
In this subsection, we demonstrate that AKD mitigates MIAs by reducing the generalization gap. Following~\cite{par-gan}, we show qualitative and quantitative evaluations.
% that applying the proposed defense reduces the generalization gap for image translation models. 

\begin{figure*}[t!] 
\begin{subfigure}{0.48\linewidth}\centering\includegraphics[width=0.7\columnwidth]{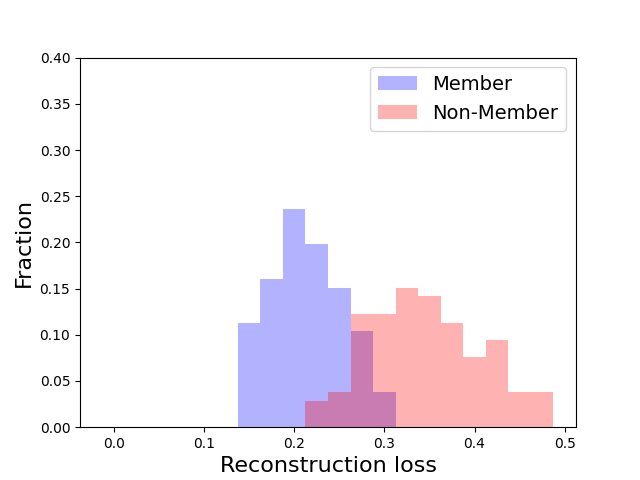}\caption{Facade, Regular training} \end{subfigure}
\begin{subfigure}{0.48\linewidth}\centering\includegraphics[width=0.7\columnwidth]{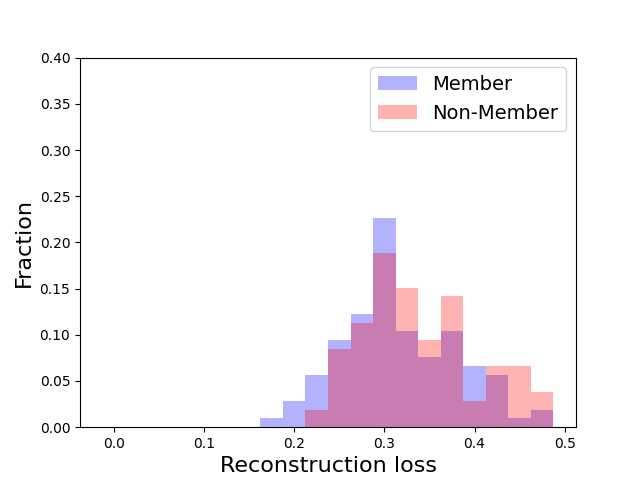}\caption{Facade, AKD}\end{subfigure} \\
\begin{subfigure}{0.48\linewidth}\centering\includegraphics[width=0.7\columnwidth]{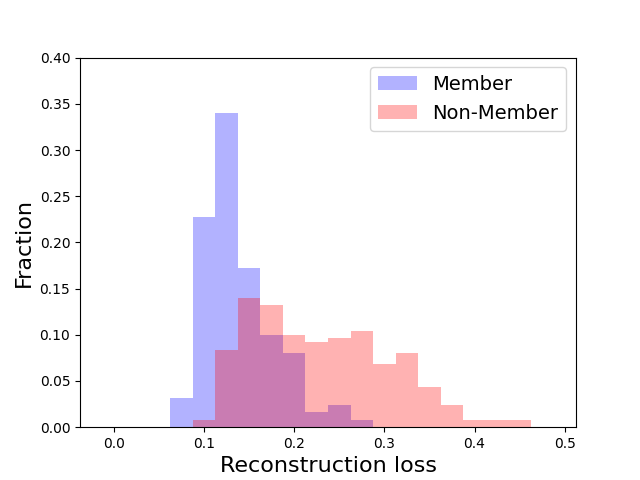}\caption{Cityscapes, Regular training}\end{subfigure}
\begin{subfigure}{0.48\linewidth}\centering\includegraphics[width=0.7\columnwidth]{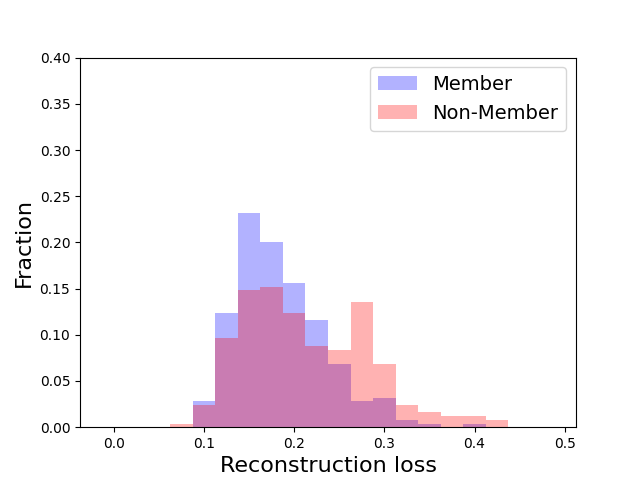}\caption{Cityscapes, AKD}\end{subfigure}
\caption{The histogram of the reconstruction loss. Top: Facade (architectural labels to real building images), Bottom: Cityscapes (semantic map to real scenes) Left: Regular training, Right: Model trained with AKD.}
\label{fig:loss_compare}
\end{figure*}

For the qualitative experiment, the reconstruction loss Eq.\eqref{eq:recon_loss} between the generated sample and the ground truth is computed for member samples (from the training set) and non-member samples (from the testing set). The histograms of the loss distributions are illustrated in Fig.~\ref{fig:loss_compare} for the regular training models and for the models trained with AKD. We observe that for regular training models, the distributions of the loss values for the member samples and non-member samples are distinct. Such distinction is the reason for easy MIAs on regular training image translation models, which is aligned with the findings in~\cite{ICCV'21}. On the other hand, for AKD, the distributions of the loss values for members and non-members are overlapping. That is to say, the behaviours of the network outputs from training samples and test samples are more similar for AKD compared to %than that of 
regular training. This shows that the model obtained by AKD generalizes well, whereas the regular training model performs differently on training and testing samples. Indeed, the better generalization improves the robustness of the model against  MIAs. 

% The comparison of the distributions also reveals the reason behind the the difference in AUCROCs of the regular training model on two tasks. The overlapping area between the member loss distribution and non-members loss distribution for the regular training model in Facade is larger in Cityspaces. Hence, the second task's model is more robust against MIA.  

%which is closely associated with overfitting to the training dataset. The generalization gap is defined as the difference between a model's performance on its training data and its performance on the unseen data from the same distribution (e,g. testing dataset). The generalization gap $g$ for image translation model $G$ is measured as:
%\begin{equation}
%\label{eq:gen_gap}
%g(G) = Q_{\mathbf{x} \in {X_{train}}} (G(\mathbf{x})) - Q_{\mathbf{x} \in {X_{test}}} (G(\mathbf{x}))
%\end{equation}

%Where $Q(.)$ measures the quality of the generated samples. NKID is used in our case to measure quality. $X_{train}=\{\mathbf{x}^{(n)}\}_{n=1}^{m}$ is taken from samples in ($S_{train}$) and $X_{test}=\{\mathbf{x}^{(n)}\}_{n=1}^{m}$ is a subset of testing dataset's inputs to the model. 

For the quantitative evaluation, the generalization gap is measured by Eq.\eqref{eq:gen_gap} for the regular training model and the model trained using AKD. The generalization gap $g$ shown in Table~\ref{tbl:gen} is computed as the NKID differences on training set and testing set for two tasks. The results in Table~\ref{tbl:gen} confirm that the generalization gap is lowered by applying the AKD defense. Therefore, overfitting is reduced using the AKD defense which is another advantage of the proposed method in addition to reduction in the risk from MIAs.

\begin{table}[t!]
\centering
\caption{ The average and standard deviation of the generalization gap for two tasks}
\label{tbl:gen}
\begin{tabular}{|c|c|}
\hline
 \multicolumn{2}{|c|}{Facade}   \\ \hline \hline
\textbf{}          & generalization gap $g$              \\ \hline
Regular   Training &  1.15 $\pm$ 0.2  \\ 
AKD                &  0.59 $\pm$ 0.23     \\ \hline \hline
 \multicolumn{2}{|c|}{Citysacpes}   \\ \hline \hline
 \textbf{}         &  generalization gap $g$              \\ \hline
Regular   Training &   0.46 $\pm$  0.24 \\ 
AKD                &   0.34 $\pm$  0.30 \\ \hline
\end{tabular}
\end{table}

%% file: Sections/06_relatedwork.tex
\section{Related Work}
\label{sec:related_work}
\subsection{Membership Inference Attacks}
% P1: general mia - 
%     S1: source of mia (refer to the survey)
%     S2: application on image
%     S3: category the methods: white-box and black-box and brief typical methods; for black-box - sub categories
MIAs has been extensively studied in some other research fields, like mobility privacy~\cite{pyrgelis2018knock}. In 2017, ~\cite{shokri2017membership} is the first work to apply MIAs against machine learning. 
Since then, there have been an increasing number of studies that investigate MIAs on various models in the computer vision domain, such as image classification~\cite{shokri2017membership,overfitting,rezaei2021difficulty}, image segmentation~\cite{he2020segmentations}, and image generation~\cite{par-gan,ICCV'21,mukherjee2021privgan}. 

As introduced in Sec.\ref{sec:prem}, based on adversarial knowledge, we can roughly categorize MIAs into two categories~\cite{nasr_S&P}: white-box attacks, in which an attacker can get {all weights of the model} and use it to attack the model, and black-box attacks, in which the attacker is given limited information on the model, but only can query it. 
Comparing with white-box attacks, black-box attacks are more dangerous because the attacker can breach the membership privacy with limited knowledge. 
%  black-box queries on the target model

More specifically, there are two major types of black-box membership inference attacks: %Based on how to construct the black-box attack models, there are two major types: 
shadow %training 
model based attacks and metric based attacks. %The first study~\cite{shokri2017membership} is a typical shadow training based attack. 
The main idea in shadow model based attacks is that an adversary can create multiple shadow models to mimic the behavior of the target model, because the adversary is assumed to know the structure and the learning algorithm of the target model.
%As 
The shadow model based attacks rely on a binary classifier to recognize the complex relationship between members and non-members. Meanwhile, metric based MIAs make membership inference decisions for data records by calculating metrics on their prediction vectors, which is simpler and requires less computational power compared to shadow model based attacks. 
The metrics used in MIAs include prediction correctness~\cite{sablayrolles2019white,overfitting}, prediction loss~\cite{sablayrolles2019white,overfitting}, prediction entropy~\cite{salem2019ml}, etc.  

% P2: mia for im2im (why not gan? but only im2im) S1: why im2im? (overfit on classification/high FAR) S1': GAN? S2: the attack method used in im2im mia S3: the target of this paper? 
%However, 
Despite many works on MIAs, a series of studies recognize MIAs are not practical. For example, a recent study~\cite{rezaei2021difficulty} shows that the high false alarm rate (i.e., predicting non-members as members)  makes MIAs fundamentally impractical in  image classification tasks.  {Similarly, it is shown in~\cite{gan-leaks} that the black-box MIA becomes less effective for Generative Adversarial Networks (GAN) trained on larger datasets, e.g. CIFAR10.} 
% Furthermore, more training data also reduces the effectiveness of MIA on generative models, like~\cite{logan}.

{Nevertheless}, image translation models are more vulnerable against MIAs as shown in~\cite{ICCV'21}.  The reasons for such vulnerability has three-fold.  Firstly, {in image translation models uncertainty in the prediction
of the output given an input is high\cite{ICCV'21}.} %the tasks where there is more uncertainty in the prediction
%of the output given an input are more susceptible to MIA.  
Secondly, {the output of the image translation models is high dimensional data~\cite{ICCV'21}.} %tasks with higher-dimensional outputs are more vulnerable to MIA.  
Lastly, {image translation datasets are generally small and the model trained with small training data is more vulnerable to MIAs~\cite{logan}}.  

%As a result, ~\cite{ICCV'21} proposed a metric based method relying on prediction errors as an effective MIA for image translation tasks, 
Authors in ~\cite{ICCV'21} also validate the ineffectiveness of existing MIA defense methods on image translation models. %To deal with the potential threats in~\cite{ICCV'21}, 
Our study focuses on an effective defense method against MIAs on image translation tasks. 

\subsection{Defense on MIAs}
% P1: general method (as DP-SGD, noise, etc.; why not regularization? why not reconstruction loss and argmax in exp?)
% P1': defense on GAN (why PrivGAN or PRAGAN are not baselines?) cost on high dimensional data?
% P2: distillation methods: description and drawbacks on applying them on im2im

%To mitigate the threatens from MIAs
%To inject 
Injecting noises into the model %training process 
including adding noises on the generated samples and differential privacy,
is widely used to mitigate the privacy threat from MIAs. %including adding noises on the samples and differential privacy.
% \saeed{Cryptography} methods, including adding noise on the samples and differential privacy, are %the most 
% widely used to mitigate the privacy threat from MIAs.  
The Gauss %ian 
defense as introduced in Sec.\ref{sec:exp_setup} is a typical method %as
of adding noise %on 
to the sample~\cite{gauss}. 
Moreover, differential privacy has been regarded as a strong privacy standard~\cite{DP-SGD}.  Thus, we also choose DP-SGD~\cite{DP-SGD} which is a classical differential privacy method, as a baseline in Sec.\ref{sec:exp_setup}.  According to  the results %from our 
in Sec.~\ref{sec:exp} as well as the results in \cite{ICCV'21}, these methods cannot mitigate the MIA in the image translation task successfully.

Knowledge distillation is an emerging thread %of defensing MIAs. 
in the defense methods against MIAs.
DMP~\cite{DMP} is the first study to apply the knowledge distillation idea on defending MIAs.  It restricts the private classifier’s direct access to the private training dataset, thus significantly reduces the membership information leakage.  
Following DMP, there are two studies SELENA~\cite{tang2021mitigating} and CKD/PCKD~\cite{zheng2021resisting} which both split the original dataset into subsets, and leverage the subset models to distill the final public model.  
The advantage of these two DMP followers is to avoid  the need for extra public data that may be hard to obtain in some applications. As mentioned in Sec.~\ref{sec:intro}, none of the existing knowledge distillation based defense method is designed for the generative models. More specifically, all of the above knowledge distillation defenses are applied to obtain the classification label/soft-label from the teacher models to prevent the final public model access the full information of the training data. However, as claimed in Sec.~\ref{sec:intro}, 
it is not trivial to apply them directly on image translation tasks. As a result, we combine adversarial training with knowledge distillation to solve the MIA issue on image translation tasks.  

%% file: Sections/07_conclusion.tex
\section{Conclusion}
\label{sec:conc}
In this paper, Adversarial Knowledge Distillation (AKD) is proposed and used as a defense method against the membership inference attack on image translation models. In the proposed method, the private generator is employed as the teacher model to distill knowledge to the student generator in an adversarial training using an unlabelled proxy dataset. 

The experimental results show that the proposed method reduces the MIA AUCROC up to 38.9\% with a minor degradation in the quality of the generated samples. The experimental results also show that the proposed method achieves the best utility-privacy tradeoff compared with the existing defenses.